\documentclass[11pt,a4paper]{article}
\PassOptionsToPackage{hyphens}{url}\usepackage[hyperref]{emnlp2018}
\usepackage{times}
\usepackage{latexsym}
\usepackage{amsmath}
\usepackage[colorinlistoftodos]{todonotes}
\usepackage{wasysym}
\usepackage[utf8]{inputenc}
\usepackage{url}
\usepackage[printwatermark]{xwatermark}

\newwatermark[pages=1,color=black!50,angle=0,scale=0.2,xpos=0,ypos=-135]{2018 EMNLP Workshop \\ on \\ Analyzing and Interpreting Neural Networks for NLP \\ (BlackboxNLP)}
\aclfinalcopy 

\title{Learning Explanations from Language Data}

\author{David Harbecke* \\
  \\\And
  Robert Schwarzenberg* \\
  German Research Center for Artificial Intelligence (DFKI) \\ Alt-Moabit 91c, 10559 Berlin, Germany\\  
  {\tt \{firstname.lastname\}@dfki.de} \\\And
   Christoph Alt \\
  }

\date{}

\begin{document}

\maketitle

\begin{abstract}
PatternAttribution is a recent method, introduced in the vision domain, that explains classifications of deep neural networks. We demonstrate that it also generates meaningful interpretations in the language domain.
\end{abstract}

\section{Introduction}

In the last decade, deep neural classifiers achieved state-of-the-art results in many domains, among others in vision and language. Due to the complexity of a deep neural model, however, it is difficult to explain its decisions. Understanding its decision process potentially allows to improve the model and may reveal new knowledge about the input. 

Recently, \citet{kindermans2018learning} claimed that ``popular explanation approaches for neural networks (...) do not provide the correct explanation, even for a simple linear model.'' They show that in a linear model, the weights serve to cancel noise in the input data and thus the weights show how to extract the signal but not what the signal is. This is why explanation methods need to move beyond the weights, the authors explain, and they propose the methods ``PatternNet'' and ``PatternAttribution'' that learn explanations from data. We test their approach in the language domain and point to room for improvement in the new framework.

\section{Methods}

\citet{kindermans2018learning} assume that the data $x$ passed to a linear model $w^{T}x = y$ is composed of signal ($s$) and noise ($d$, from distraction) $x = s + d$. Furthermore, they also assume that there is a linear relation between signal and target $ya_{s} = s$ where $a_{s}$ is a so called signal base vector, which is in fact the ``pattern'' that PatternNet finds for us. As mentioned in the introduction, the authors show that in the model above, $w$ serves to cancel the noise such that 
\begin{equation}\label{eq:simpleconditions}
w^{T}d = 0,\quad w^{T}s = y.
\end{equation}
They go on to explain that a good signal estimator $S(x)=\hat{s}$ should comply to the conditions in Eqs.~\ref{eq:simpleconditions} but that these alone form an ill-posed quality criterion since $S(x) = u(w^{T}u)^{-1}y $ already satisfies them for any $u$ for which $w^{T}u\neq0$. To address this issue they introduce another quality criterion over a batch of data $\mathbf{x}$:
\begin{equation}\label{eq:criterion}
\rho(S) = 1 - \max_{v} \operatorname{corr}(\overbrace{w^{T}\mathbf{x}}^{\mathbf{y}}, v^{T}\overbrace{(\mathbf{x}-S(\mathbf{x}))}^{\hat{\mathbf{d}}})
\end{equation}
and point out that Eq.~\ref{eq:criterion} yields maximum values for signal estimators that remove most of the information about $\mathbf{y}$ in the noise.

We argue that Eq.~\ref{eq:criterion} still is not exhaustive. Consider the artificial estimator
\begin{equation*}
S_{m}(x) = mx + (1-m)s = s + md
\end{equation*}
which arguably is a a bad signal estimator for large $m$ as its estimation contains scaled noise, $md$. Nevertheless, it still satisfies Eqs.~\ref{eq:simpleconditions} and yields maximum values for Eq.~\ref{eq:criterion} since
\begin{equation*}
x-S_{m}(x) = (1-m)(x-s) = (1-m)d
\end{equation*}
is again just scaled noise and thus does not correlate with the output $y$. To solve this issue, we propose the following criterion:
\begin{equation*}
\begin{split}
\rho' (S) :=& \max_{v_{1}} \operatorname{corr}(w^{T}\mathbf{x}, v_{1}^{T}S(\mathbf{x})) \\ &- \max_{v_{2}} \operatorname{corr}(w^{T}\mathbf{x}, v_{2}^{T}(\mathbf{x}-S(\mathbf{x}))).
\end{split}
\end{equation*}
The minuend measures how much noise is left in the signal, the subtrahend measures how much signal is left in the noise. Good signal estimators split signal and noise well and thus yield large $\rho' (S)$. We leave it to future research to evaluate existing signal estimators with our new criterion.

\begin{figure}[!htb] 
\small
    \colorbox[RGB]{255,140,140}{\strut Great} \colorbox[RGB]{255,175,175}{\strut book} \colorbox[RGB]{255,237,237}{\strut for} \colorbox[RGB]{255,234,234}{\strut travelling} \colorbox[RGB]{255,227,227}{\strut Europe} \colorbox[RGB]{255,237,237}{\strut :} \colorbox[RGB]{255,255,255}{\strut I} \colorbox[RGB]{255,239,239}{\strut currently} \colorbox[RGB]{255,228,228}{\strut live} \colorbox[RGB]{255,240,240}{\strut in} \colorbox[RGB]{255,247,247}{\strut Europe} \colorbox[RGB]{255,252,252}{\strut ,} \colorbox[RGB]{255,252,252}{\strut and} \colorbox[RGB]{255,255,255}{\strut this} \colorbox[RGB]{255,244,244}{\strut is} \colorbox[RGB]{255,247,247}{\strut the} \colorbox[RGB]{255,255,255}{\strut book} \colorbox[RGB]{255,183,183}{\strut I} \colorbox[RGB]{255,183,183}{\strut recommend} \colorbox[RGB]{255,248,248}{\strut for} \colorbox[RGB]{255,248,248}{\strut my} \colorbox[RGB]{255,244,244}{\strut visitors} \colorbox[RGB]{255,252,252}{\strut .} \colorbox[RGB]{255,245,245}{\strut It} \colorbox[RGB]{255,245,245}{\strut covers} \colorbox[RGB]{255,230,230}{\strut many} \colorbox[RGB]{255,217,217}{\strut countries} \colorbox[RGB]{255,224,224}{\strut ,} \colorbox[RGB]{255,230,230}{\strut colour} \colorbox[RGB]{255,252,252}{\strut pictures} \colorbox[RGB]{255,255,255}{\strut ,} \colorbox[RGB]{255,250,250}{\strut and} \colorbox[RGB]{255,248,248}{\strut is} \colorbox[RGB]{255,247,247}{\strut a} \colorbox[RGB]{255,177,177}{\strut nice} \colorbox[RGB]{255,168,168}{\strut starter} \colorbox[RGB]{255,244,244}{\strut for} \colorbox[RGB]{255,255,255}{\strut before} \colorbox[RGB]{255,255,255}{\strut you} \colorbox[RGB]{255,255,255}{\strut go} \colorbox[RGB]{255,255,255}{\strut ,} \colorbox[RGB]{255,255,255}{\strut and} \colorbox[RGB]{255,255,255}{\strut once} \colorbox[RGB]{255,255,255}{\strut you} \colorbox[RGB]{255,255,255}{\strut are} \colorbox[RGB]{255,255,255}{\strut there}\colorbox[RGB]{255,255,255}{\strut .}
        \caption{Contributions to positive classification.}
        \label{fig:example_sentence_1}
\end{figure}

For our experiments, the authors equip us with expressions for the signal base vectors $a_{s}$ for simple linear layers and ReLU layers. For the simple linear model, for instance, it turns out that $a_{s} = \operatorname{cov}(\mathbf{x},\mathbf{y})/ \sigma_{\mathbf{y}}^{2}$. To retrieve contributions for PatternAttribution, in the backward pass, the authors replace the weights by $w \astrosun a_{s}$.

\section{Experiments}

To test PatternAttribution in the NLP domain, we trained a CNN text classifier \cite{CNNTextClassifier} on a subset of the Amazon review polarity data set \cite{zhang2015character}. We used 150 bigram filters, dropout regularization and a dense FC projection with 128 neurons. Our classifier achieves an F$_{1}$ score of 0.875 on a fixed test split. We then used \citet{kindermans2018learning} PatternAttribution to retrieve neuron-wise signal contributions in the input vector space.\footnote{Our experiments are available at \url{https://github.com/DFKI-NLP/language-attributions}.}

To align these contributions with plain text, we summed up the contribution scores over the word vector dimensions for each word and used the accumulated scores to scale RGB values for word highlights in the plain text space. Positive scores are highlighted in red, negative scores in blue. This approach is inspired by \citet{arras-plos17}. Example contributions are shown in Figs.~\ref{fig:example_sentence_1} and \ref{fig:example_sentence_2}.

\begin{figure}[!htb] 
\small
\colorbox[RGB]{240,240,255}{\strut DVD} \colorbox[RGB]{224,224,255}{\strut Player} \colorbox[RGB]{205,205,255}{\strut crapped} \colorbox[RGB]{228,228,255}{\strut out} \colorbox[RGB]{255,255,255}{\strut after} \colorbox[RGB]{254,254,255}{\strut one} \colorbox[RGB]{253,253,255}{\strut year} \colorbox[RGB]{254,254,255}{\strut :} \colorbox[RGB]{254,254,255}{\strut I} \colorbox[RGB]{253,253,255}{\strut also} \colorbox[RGB]{252,252,255}{\strut began} \colorbox[RGB]{255,255,255}{\strut having} \colorbox[RGB]{238,238,255}{\strut the} \colorbox[RGB]{167,167,255}{\strut incorrect} \colorbox[RGB]{192,192,255}{\strut disc} \colorbox[RGB]{247,247,255}{\strut problems} \colorbox[RGB]{255,255,255}{\strut that} \colorbox[RGB]{255,255,255}{\strut I} \colorbox[RGB]{246,246,255}{\strut 've} \colorbox[RGB]{243,243,255}{\strut read} \colorbox[RGB]{255,255,255}{\strut about} \colorbox[RGB]{254,254,255}{\strut on} \colorbox[RGB]{254,254,255}{\strut here} \colorbox[RGB]{255,255,255}{\strut .} \colorbox[RGB]{253,253,255}{\strut The} \colorbox[RGB]{248,248,255}{\strut VCR} \colorbox[RGB]{244,244,255}{\strut still} \colorbox[RGB]{240,240,255}{\strut works} \colorbox[RGB]{249,249,255}{\strut ,} \colorbox[RGB]{255,255,255}{\strut but} \colorbox[RGB]{212,212,255}{\strut hte} \colorbox[RGB]{205,205,255}{\strut DVD} \colorbox[RGB]{249,249,255}{\strut side} \colorbox[RGB]{175,175,255}{\strut is} \colorbox[RGB]{149,149,255}{\strut useless} \colorbox[RGB]{221,221,255}{\strut .} \colorbox[RGB]{255,255,255}{\strut I} \colorbox[RGB]{255,255,255}{\strut understand} \colorbox[RGB]{255,255,255}{\strut that} \colorbox[RGB]{241,241,255}{\strut DVD} \colorbox[RGB]{246,246,255}{\strut players} \colorbox[RGB]{255,255,255}{\strut sometimes} \colorbox[RGB]{234,234,255}{\strut just} \colorbox[RGB]{236,236,255}{\strut quit} \colorbox[RGB]{255,255,255}{\strut on} \colorbox[RGB]{255,255,255}{\strut you} \colorbox[RGB]{255,255,255}{\strut ,} \colorbox[RGB]{255,255,255}{\strut but} \colorbox[RGB]{239,239,255}{\strut after} \colorbox[RGB]{242,242,255}{\strut not} \colorbox[RGB]{255,255,255}{\strut even} \colorbox[RGB]{255,255,255}{\strut one} \colorbox[RGB]{238,238,255}{\strut year} \colorbox[RGB]{244,244,255}{\strut ?} \colorbox[RGB]{227,227,255}{\strut To} \colorbox[RGB]{245,245,255}{\strut me} \colorbox[RGB]{255,255,255}{\strut that} \colorbox[RGB]{251,251,255}{\strut 's} \colorbox[RGB]{248,248,255}{\strut a} \colorbox[RGB]{252,252,255}{\strut sign} \colorbox[RGB]{244,244,255}{\strut on} \colorbox[RGB]{112,112,255}{\strut bad} \colorbox[RGB]{144,144,255}{\strut quality} \colorbox[RGB]{248,248,255}{\strut .} \colorbox[RGB]{255,255,255}{\strut I} \colorbox[RGB]{255,255,255}{\strut 'm} \colorbox[RGB]{255,255,255}{\strut giving} \colorbox[RGB]{250,250,255}{\strut up} \colorbox[RGB]{247,247,255}{\strut JVC} \colorbox[RGB]{250,250,255}{\strut after} \colorbox[RGB]{253,253,255}{\strut this} \colorbox[RGB]{255,255,255}{\strut as} \colorbox[RGB]{255,255,255}{\strut well} \colorbox[RGB]{255,255,255}{\strut .} \colorbox[RGB]{255,255,255}{\strut I} \colorbox[RGB]{248,248,255}{\strut 'm} \colorbox[RGB]{233,233,255}{\strut sticking} \colorbox[RGB]{235,235,255}{\strut to} \colorbox[RGB]{252,252,255}{\strut Sony} \colorbox[RGB]{255,255,255}{\strut or} \colorbox[RGB]{255,255,255}{\strut giving} \colorbox[RGB]{251,251,255}{\strut another} \colorbox[RGB]{248,248,255}{\strut brand} \colorbox[RGB]{252,252,255}{\strut a} \colorbox[RGB]{255,255,255}{\strut shot}\colorbox[RGB]{255,255,255}{\strut .}
        \caption{Contributions to negative classification.}
        \label{fig:example_sentence_2}
\end{figure}

\section{Results}

We observe that bigrams are highlighted, in particular no highlighted token stands isolated. Bigrams with clear positive or negative sentiment contribute heavily to the sentiment classification. In contrast, stop words and uninformative bigrams make little to no contribution. We consider these meaningful explanations of the sentiment classifications. 

\section{Related Work}

Many of the approaches used to explain and interpret models in NLP mirror methods originally developed in the vision domain, such as the recent approaches by \citet{li2016visualizing}, \citet{arras-plos17}, and \citet{arras2017explaining}. In this paper we implemented a similar strategy. 

Following \citet{kindermans2018learning}, however, our approach improves upon the latter methods for the reasons outlined above. Furthermore, PatternAttribution is related to \citet{montavon2017explaining} who make use of Taylor decompositions to explain deep models. PatternAttribution reveals a good root point for the decomposition, the authors explain. 

\section{Conclusion}
We successfully transferred a new explanation method to the NLP domain. We were able to demonstrate that PatternAttribution can be used to identify meaningful signal contributions in text inputs. Our method should be extended to other popular models in NLP. Furthermore, we introduced an improved quality criterion for signal estimators. In the future, estimators can be deduced from and tested against our new criterion.
\let\thefootnote\relax\footnotetext{* Co-first authorship.}\let\thefootnote\relax\footnotetext{This  research  was  partially  supported  by the German Federal Ministry of Education and Research through the projects DEEPLEE (01IW17001) and BBDC (01IS14013E).}

\bibliography{emnlp2018}

\begin{thebibliography}{7}
\expandafter\ifx\csname natexlab\endcsname\relax\def\natexlab#1{#1}\fi

\bibitem[{Arras et~al.(2017{\natexlab{a}})Arras, Horn, Montavon, M{\"u}ller,
  and Samek}]{arras-plos17}
Leila Arras, Franziska Horn, Gr{\'e}goire Montavon, Klaus-Robert M{\"u}ller,
  and Wojciech Samek. 2017{\natexlab{a}}.
\newblock "{W}hat is relevant in a text document?": An interpretable machine
  learning approach.
\newblock \emph{PLOS ONE}, 12(8).

\bibitem[{Arras et~al.(2017{\natexlab{b}})Arras, Montavon, M{\"u}ller, and
  Samek}]{arras2017explaining}
Leila Arras, Gr{\'e}goire Montavon, Klaus-Robert M{\"u}ller, and Wojciech
  Samek. 2017{\natexlab{b}}.
\newblock Explaining recurrent neural network predictions in sentiment
  analysis.
\newblock In \emph{Proceedings of the 8th EMNLP Workshop on Computational
  Approaches to Subjectivity, Sentiment and Social Media Analysis}, pages
  159--168.

\bibitem[{Kim(2014)}]{CNNTextClassifier}
Yoon Kim. 2014.
\newblock Convolutional neural networks for sentence classification.
\newblock In \emph{Proceedings of the 2014 Conference on Empirical Methods in
  Natural Language Processing (EMNLP)}, pages 1746--1751.

\bibitem[{Kindermans et~al.(2018)Kindermans, Schütt, Alber, Müller, Erhan,
  Kim, and Dähne}]{kindermans2018learning}
Pieter-Jan Kindermans, Kristof~T. Schütt, Maximilian Alber, Klaus-Robert
  Müller, Dumitru Erhan, Been Kim, and Sven Dähne. 2018.
\newblock Learning how to explain neural networks: {PatternNet} and
  {PatternAttribution}.
\newblock In \emph{International Conference on Learning Representations
  (ICLR)}.

\bibitem[{Li et~al.(2016)Li, Chen, Hovy, and Jurafsky}]{li2016visualizing}
Jiwei Li, Xinlei Chen, Eduard Hovy, and Dan Jurafsky. 2016.
\newblock Visualizing and understanding neural models in {NLP}.
\newblock In \emph{Proceedings of NAACL-HLT}, pages 681--691.

\bibitem[{Montavon et~al.(2017)Montavon, Lapuschkin, Binder, Samek, and
  M{\"u}ller}]{montavon2017explaining}
Gr{\'e}goire Montavon, Sebastian Lapuschkin, Alexander Binder, Wojciech Samek,
  and Klaus-Robert M{\"u}ller. 2017.
\newblock Explaining nonlinear classification decisions with deep taylor
  decomposition.
\newblock \emph{Pattern Recognition}, 65:211--222.

\bibitem[{Zhang et~al.(2015)Zhang, Zhao, and LeCun}]{zhang2015character}
Xiang Zhang, Junbo Zhao, and Yann LeCun. 2015.
\newblock Character-level convolutional networks for text classification.
\newblock In \emph{Advances in Neural Information Processing Systems (NIPS)},
  pages 649--657.

\end{thebibliography}
\bibliographystyle{acl_natbib_nourl}

\end{document}